\documentclass{article}

\usepackage[nonatbib,preprint]{neurips_2025}
\usepackage{cite}
\usepackage{amsmath,amssymb}
\usepackage{algorithmic}
\usepackage{graphicx}
\usepackage{textcomp}
\usepackage{xcolor}
\usepackage{tikz}
\usepackage{adjustbox}
\usepackage{multirow}
\usepackage{array,tabularx,booktabs}
\usepackage{siunitx}
\usepackage{makecell}
\usepackage{threeparttable}
\usetikzlibrary{positioning,backgrounds,calc,arrows}
\usepackage{hyperref}
\usepackage{subcaption}
\usetikzlibrary{shapes.geometric}
\usepackage{xspace}
\usepackage[utf8]{inputenc}
\usepackage[T1]{fontenc}

\newcommand{\tsn}[1]{{\left\vert\kern-0.25ex\left\vert\kern-0.25ex\left\vert #1 
    \right\vert\kern-0.25ex\right\vert\kern-0.25ex\right\vert}}
\usepackage{xcolor}
\definecolor{darkred}{RGB}{150,0,0}
\definecolor{darkgreen}{RGB}{0,150,0}
\definecolor{darkblue}{RGB}{0,0,200}

\newcommand{\algname}{VQ-GAN\xspace}

\newcommand{\alg}{\textsc{PureVQ-GAN}\xspace}

\newcommand{\beq}{\begin{equation}}

\newcommand{\eeq}{\end{equation}}

\def \endprf{\hfill {\vrule height6pt width6pt depth0pt}\medskip}

\title{\alg: Defending Data Poisoning Attacks through Vector-Quantized Bottlenecks}

\author{%
  Alexander Branch$^{1*}$ \quad
  Omead Pooladzandi$^{2*}$ \quad
  Radin Khosraviani$^{*}$ \\
  Sunay Gajanan Bhat$^{1}$ \quad
  Jeffrey Jiang$^{1}$ \quad
  Gregory Pottie$^{1}$ \\[0.5em]
  $^{1}$University of California, Los Angeles \quad
  $^{2}$California Institute of Technology \\
  $^{*}$Equal contribution \\[0.3em]
  \texttt{\{alexrbranch, sunaybhat1, jimmery\}@ucla.edu, pottie@ee.ucla.edu} \\
  \texttt{omead@caltech.edu, radin.msft@gmail.com}
}

\begin{document}
\raggedbottom  

\maketitle

\begin{abstract}
We introduce \textbf{\alg}, a defense against data poisoning that forces backdoor triggers through a discrete bottleneck using Vector-Quantized VAE with GAN discriminator. By quantizing poisoned images through a learned codebook, \alg\ destroys fine-grained trigger patterns while preserving semantic content. A GAN discriminator ensures outputs match the natural image distribution, preventing reconstruction of out-of-distribution perturbations. On CIFAR-10, \alg\ achieves 0\% poison success rate (PSR) against Gradient Matching and Bullseye Polytope attacks, and 1.64\% against Narcissus—while maintaining 91-95\% clean accuracy. Unlike diffusion-based defenses requiring hundreds of iterative refinement steps, \alg\ is over 50× faster, making it practical for real training pipelines.
\end{abstract}
\vspace{-0.5cm}  
\nopagebreak[4]  
\section{Introduction}

Data poisoning attacks compromise machine learning models by injecting malicious training samples that cause targeted misclassification~\cite{ding2019trojan, gu2019badnetsidentifyingvulnerabilitiesmachine}. Backdoor attacks are a critical subclass where models behave normally on clean inputs but produce attacker-chosen outputs when a trigger is present. Clean-label attacks~\cite{turner2018clean} are especially insidious - the poisoned samples are correctly labeled and visually indistinguishable from clean data, evading human inspection and statistical filtering.

\textbf{Attack Mechanisms:} Data poisoning attacks inject $\ell_\infty$-bounded adversarial perturbations $\delta$ with $\|\delta\|_\infty \leq \xi$ (typically $\xi = 8/255$ or $16/255$) into a small fraction $\alpha$ (typically 1\%) of training data. Both attack categories operate within this constraint to maintain imperceptibility:

\textit{Triggered attacks} embed a specific trigger $\rho$ with $\|\rho\|_\infty \leq \xi$ that causes misclassification to target label $y^{\text{adv}}$ when present at inference. Narcissus~\cite{zeng2022narcissuspracticalcleanlabelbackdoor} exemplifies clean-label backdoor attacks where poisoned samples appear visually normal.

\textit{Triggerless attacks} add perturbations $\epsilon$ with $\|\epsilon\|_\infty \leq \xi$ to align poisoned samples' feature representations with target images. Gradient Matching (GM)~\cite{geiping2020witches} optimizes poisons to match target gradients during training, while Bullseye Polytope (BP)~\cite{aghakhani2021bullseye} uses a fine-tuned generator to cluster poisons around targets in feature space. These attacks achieve targeted misclassification without requiring triggers at inference time.

Current defenses have fundamental limitations. Outlier detection methods~\cite{kyaw2024epic} fail against clean-label attacks where poisons blend with normal data distributions. Certified defenses~\cite{liu2022friendly} provide theoretical guarantees but sacrifice significant accuracy (>10\% drop). Compression-based methods like JPEG~\cite{wallace1991jpeg} and quantization degrade image quality, harming downstream task performance. Recent generative purifiers~\cite{pooladzandi2024puregen} show promise but rely on expensive iterative sampling - requiring hundreds of iterative refinement steps per image.

We propose \textbf{\alg}, a novel defense that leverages the discrete bottleneck of Vector-Quantized VAEs~\cite{oord2018neuraldiscreterepresentationlearning} combined with adversarial training~\cite{esser2021taming}. Our key insight: forcing representations through a finite codebook of learned prototypes destroys adversarial perturbations while preserving semantic content. Unlike continuous autoencoders that can propagate subtle poison signals, our discrete quantization creates a hard information bottleneck that attackers cannot bypass.

\textbf{Main Contributions:} (1) \textbf{Efficient purification:} \alg\ is over 50× faster than iterative generative defenses. (2) \textbf{State-of-the-art results:} 0\% PSR against GM/BP, 1.64\% against Narcissus, while maintaining 91-95\% clean accuracy on CIFAR-10. (3) \textbf{Scalable architecture:} Demonstrated with models scaling to 300M parameters.
\section{Related Work - Generative Purifiers}

This work directly builds off of the PureGen \cite{pooladzandi2024puregen} framework. PureGen proposes training and then sampling from a generative model to purify train time perturbation based attacks. Previously, PureGen proposed initializing an MCMC chain of an EBM or diffusion model with a poisoned image and sampling around the distribution of the learned image space until the poisons are erased. 

While EBMs and Diffusion models are strong purifiers, they are computationally expensive. EBMs require gradients at inference time, and Diffusion models demand hundreds of iterative denoising steps, making both methods impractical for large-scale training pipelines.

In this paper, as an alternative to EBMs and Diffusion models, we utilize a \algname ~\cite{esser2021taming} for defense. We hypothesize this works well, as the discrete bottleneck of a VQ-VAE~\cite{oord2018neuraldiscreterepresentationlearning} filters out poison perturbations, while the GAN loss maintains sharp, realistic outputs. To our knowledge, this is the first work to successfully apply a VQ-GAN as an efficient, high-fidelity poison purifier.
\section{Method: \alg}

\subsection{Architecture}
\alg\ combines a Vector-Quantized Variational Autoencoder (VQ-VAE) generator $G$ with an adversarial discriminator $D_{\text{adv}}$. The encoder $E$ maps input $x \in \mathbb{R}^{H \times W \times 3}$ to continuous latent representation $z_e(x) \in \mathbb{R}^{h \times w \times d}$. Each spatial location in $z_e$ is quantized to its nearest codebook vector:
\begin{equation}
e_q = \arg\min_{e_k \in \{e_1, ..., e_K\}} \|z_e(x) - e_k\|_2
\end{equation}
where codebook $\{e_k\}_{k=1}^K$ contains $K$ learnable $d$-dimensional vectors. The decoder $D$ reconstructs $\hat{x} = D(e_q)$ from quantized latents. This discrete bottleneck is critical: poison perturbations cannot precisely control which codebook vectors are selected, breaking the attack.

\begin{figure}[t]
    \centering
\includegraphics[width=0.7\textwidth]{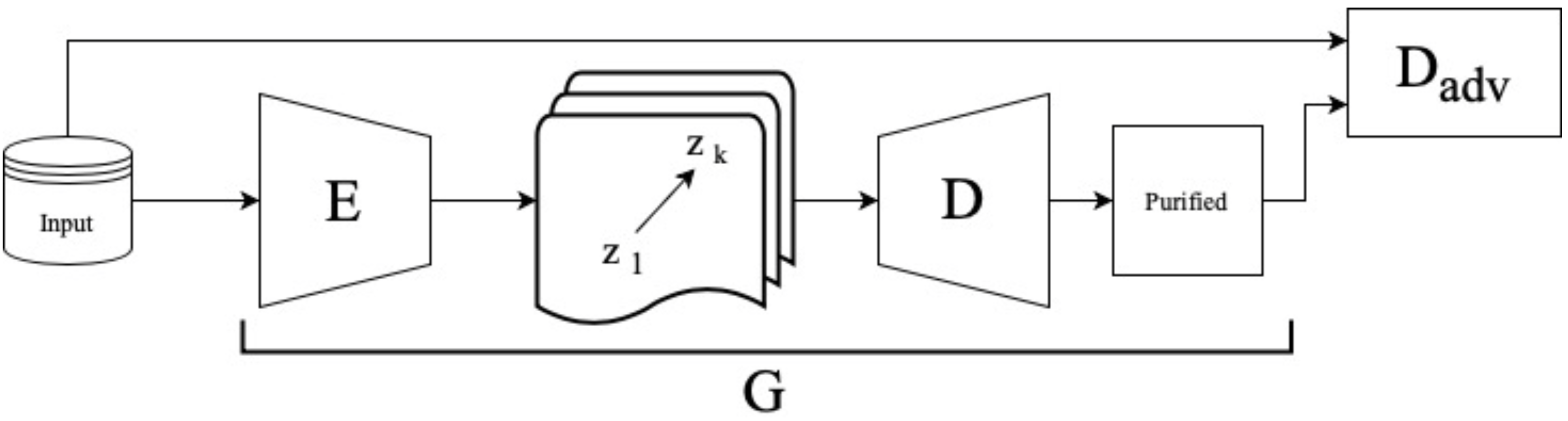}
\caption{\alg\ architecture.}
\label{fig:arch}
\end{figure}

\subsection{Training Objective}
We train \alg\ with three loss components:

\textbf{Reconstruction Loss:} $\mathcal{L}_{\text{rec}} = \|x - \hat{x}\|_2^2$ ensures faithful reconstruction of clean images.

\textbf{Codebook Learning:} Following~\cite{oord2018neuraldiscreterepresentationlearning}, we use:
\begin{equation}
\mathcal{L}_{\text{VQ}} = \beta\|z_e - \text{sg}[e_q]\|_2^2 + \|\text{sg}[z_e] - e_q\|_2^2
\end{equation}
where $\text{sg}[\cdot]$ is stop-gradient. The first term (commitment loss with $\beta = 0.25$) encourages encoder outputs to stay close to codebook vectors. The second term updates codebook vectors.

\textbf{Adversarial Loss:} To maintain perceptual quality, we add a discriminator:
\begin{equation}
\mathcal{L}_{\text{GAN}} = \mathbb{E}_x[\log D_{\text{adv}}(x)] + \mathbb{E}_x[\log(1 - D_{\text{adv}}(\hat{x}))]
\end{equation}

\textbf{Total Generator Loss:} $\mathcal{L}_G = \mathcal{L}_{\text{rec}} + \mathcal{L}_{\text{VQ}} + \lambda\mathcal{L}_{\text{GAN}}$ where $\lambda$ controls the balance between adversarial strength and accuracy. We set $\lambda = 0.1$ in our experiments.

\textbf{Implementation Details:} Codebook size $K = 512$ balances expressiveness and regularization. Full training hyperparameters and architecture details are provided in Appendix~\ref{app:implementation}.

\subsection{Defense Deployment}

\textbf{Training Phase:} Train \alg\ on the training dataset, which may contain poisoned samples (we use the same poisoned training set in our experiments). The GAN discriminator learns to distinguish real training images from reconstructions, effectively learning the natural image distribution from the overwhelming majority (99\%) of clean samples. During classifier training, pass each mini-batch through \alg\ before computing gradients.

\textbf{Inference Phase:} Purify test images before classification using a single forward pass. For potentially stronger attacks, multiple refinement passes can be applied:
\begin{equation}
x^{(t+1)} = G(x^{(t)}), \quad x^{(0)} = x_{\text{poisoned}}
\end{equation}
though our experiments show one pass is sufficient for state-of-the-art defense.

\textbf{Theoretical Insight:} The discrete bottleneck creates an information bottleneck~\cite{tishby2000information} that acts as a filter - sufficient for natural images but insufficient for encoding both image content and poison signal. Critically, the GAN discriminator provides robustness to poisoned training data: since poisoned samples constitute only a small fraction (typically 1\%) of the training set, the discriminator learns the natural image distribution from the overwhelming majority of clean samples. During reconstruction, the adversarial loss enforces that outputs match this learned clean distribution, effectively rejecting adversarial perturbations even if they were present in the training data. The discrete codebook quantization combined with the distributional constraint from the GAN creates a dual defense mechanism that purifies poisoned inputs.
\section{Experiments}

We evaluate \alg\ on CIFAR-10 against three state-of-the-art clean-label attacks:

\textbf{Attack Setup:} (1) \textbf{Gradient Matching (GM)}~\cite{geiping2020witches} - a clean-label attack where poisons are optimized to match the gradient of a target image during training. (2) \textbf{Narcissus (NS)}~\cite{zeng2022narcissuspracticalcleanlabelbackdoor} - a backdoor attack introducing a specific patterned trigger designed to be stealthy. (3) \textbf{Bullseye Polytope (BP)}~\cite{aghakhani2021bullseye} - a powerful clean-label backdoor where a fine-tuned generator produces poisons that cluster around a target image in feature space. Each attack poisons 1\% of training data.

\textbf{Training Protocol:} We train ResNet-18 classifiers on purified datasets. \alg\ uses encoder-decoder with codebook size K=512, trained on the same poisoned CIFAR-10 dataset (containing 1\% poisoned samples) used for classifier training—matching the PureGen-EBM/Diffusion training protocol. All results use a single forward pass through the purifier. Main results (Table 1) use our 300M parameter model.

\textbf{Main Results:} \alg\ achieves near-perfect defense. Against GM and BP, we achieve 0\% PSR while maintaining >91\% clean accuracy. For the harder NS attack, PSR is only 1.64\% with 94.61\% clean accuracy - the highest among defenses. 

\textbf{Baseline Comparison:} EPIC~\cite{li2024epic} shows limited success with PSR ranging from 10-42\% and degrades accuracy to as low as 81.95\%. JPEG achieves strong defense (0-1.78\% PSR) with accuracy around 90-93\%. PureGen-EBM/Diffusion achieve comparable PSR but require hundreds of iterative refinement steps. FRIENDS~\cite{chen2024amplifyingmembershipexposurecertified} reduces PSR to 0\% for GM and 8\% for BP, with 8.32\% against NS. See Appendix~\ref{app:details} for detailed qualitative comparison.

\begin{table}[ht]
\centering
\small
\setlength{\tabcolsep}{5pt}
\renewcommand{\arraystretch}{1.15}
\begin{tabularx}{\linewidth}{l *{6}{c}}
\toprule
& \multicolumn{2}{c}{\textbf{GM (1\%)}} 
& \multicolumn{2}{c}{\textbf{NS (1\%)}} 
& \multicolumn{2}{c}{\textbf{BP (Fine-Tune)}} \\
\cmidrule(lr){2-3}\cmidrule(lr){4-5}\cmidrule(lr){6-7}
\makecell[l]{\textbf{Defense}} 
& \makecell{Poison (\%) \\ $\downarrow$} 
& \makecell{Nat. Acc (\%) \\ $\uparrow$}
& \makecell{Poison (\%) \\ $\downarrow$} 
& \makecell{Nat. Acc (\%) \\ $\uparrow$}
& \makecell{Poison (\%) \\ $\downarrow$} 
& \makecell{Nat. Acc (\%) \\ $\uparrow$} \\
\midrule
None (Baseline)      & 44.00 & 94.84$_{0.20}$ & 43.95$_{33.60}$ & 94.89$_{0.20}$ & 46.00 & 89.84$_{0.90}$ \\
EPIC                 & 10.00 & 85.14$_{1.20}$ & 27.31$_{34.00}$ & 82.20$_{1.10}$ & 42.00 & 81.95$_{5.60}$ \\
FRIENDS              & \textbf{0.00} & 91.15$_{0.40}$ & 8.32$_{22.30}$ & 91.01$_{0.40}$ & 8.00 & 87.82$_{1.20}$ \\
JPEG                 & \textbf{0.00} & 90.00$_{0.19}$ & \textbf{1.78}$_{1.17}$ & 92.94$_{0.15}$ & \textbf{0.00} & 90.40$_{0.44}$ \\
PureGen-DDPM         & \textbf{0.00} & 90.93$_{0.20}$ & \textbf{1.64}$_{0.82}$ & 90.99$_{0.22}$ & \textbf{0.00} & \textbf{91.53}$_{0.15}$ \\
PureGen-EBM          & 1.00 & \textbf{92.98}$_{0.20}$ & \textbf{1.39}$_{0.80}$ & 92.92$_{0.20}$ & \textbf{0.00} & 87.52$_{1.20}$ \\
\textbf{\alg} & \textbf{0.00} & \textbf{92.80}$_{0.18}$ & \textbf{1.64}$_{0.91}$ & \textbf{94.61}$_{0.21}$ & \textbf{0.00} & \textbf{91.43}$_{0.19}$ \\
\bottomrule
\end{tabularx}
\caption{CIFAR-10 (ResNet-18) results. Poison success (↓) and natural accuracy (↑). Subscripts show standard deviation.}
\end{table}

\textbf{Computational Efficiency:} \alg\ requires only a single (or few) forward passes for purification, making it over 50× faster than iterative generative defenses that require hundreds of iterative refinement steps per image.

\begin{figure}[t]
\centering
\begin{subfigure}[b]{0.45\textwidth}
\includegraphics[width=\textwidth]{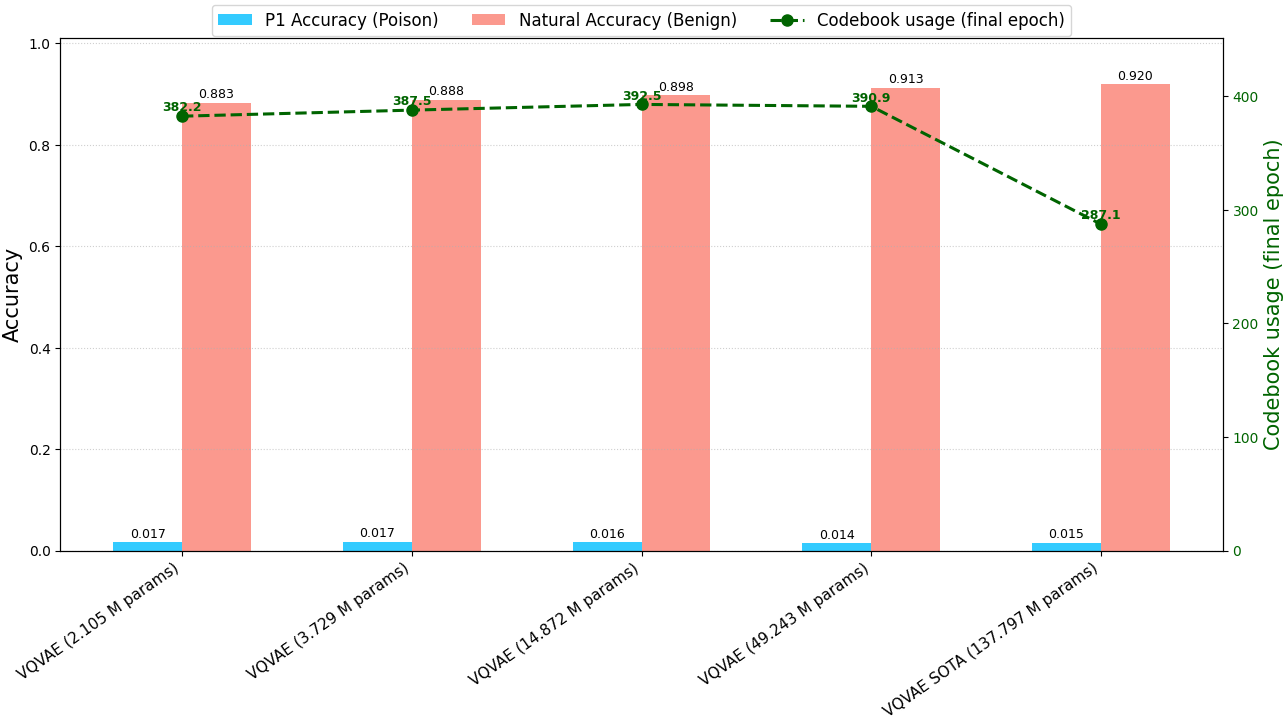}
\caption{Model size vs performance}
\end{subfigure}
\begin{subfigure}[b]{0.45\textwidth}
\includegraphics[width=\textwidth]{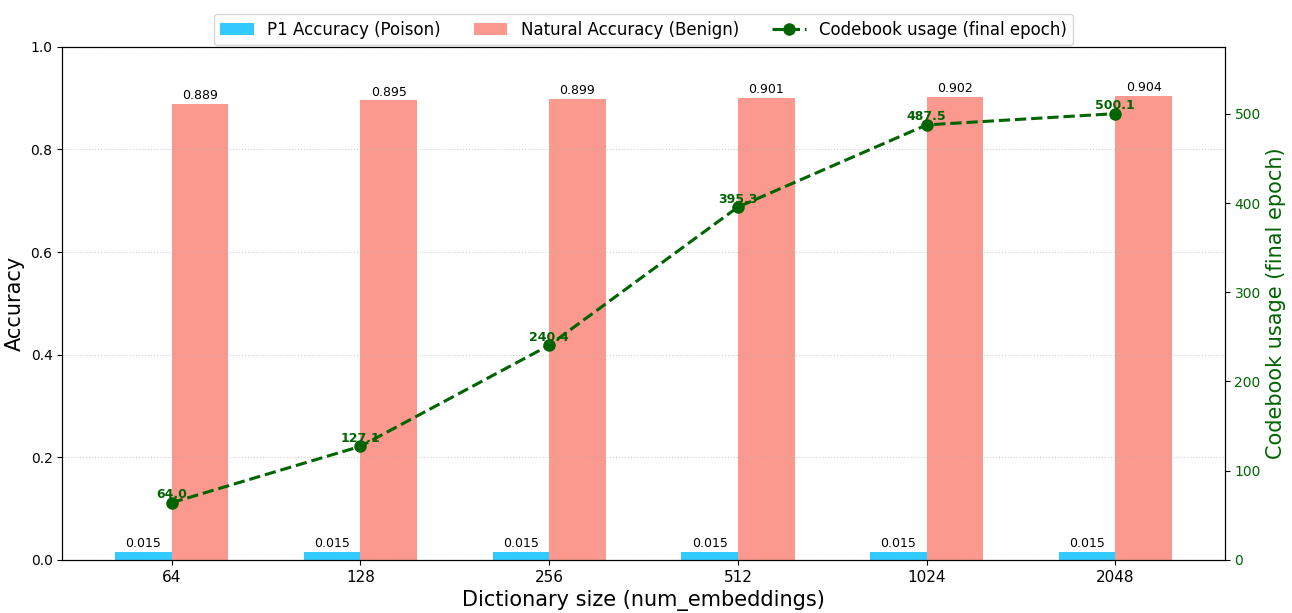}
\caption{Codebook size analysis}
\end{subfigure}
\caption{Ablations. (a) Larger models improve clean accuracy with diminishing returns >50M params (trend verified up to 300M). (b) Even K=64 achieves 0\% PSR; larger K improves reconstruction.}
\label{fig:ablation}
\end{figure}

\begin{figure}[t!]
\centering
\includegraphics[width=0.65\textwidth]{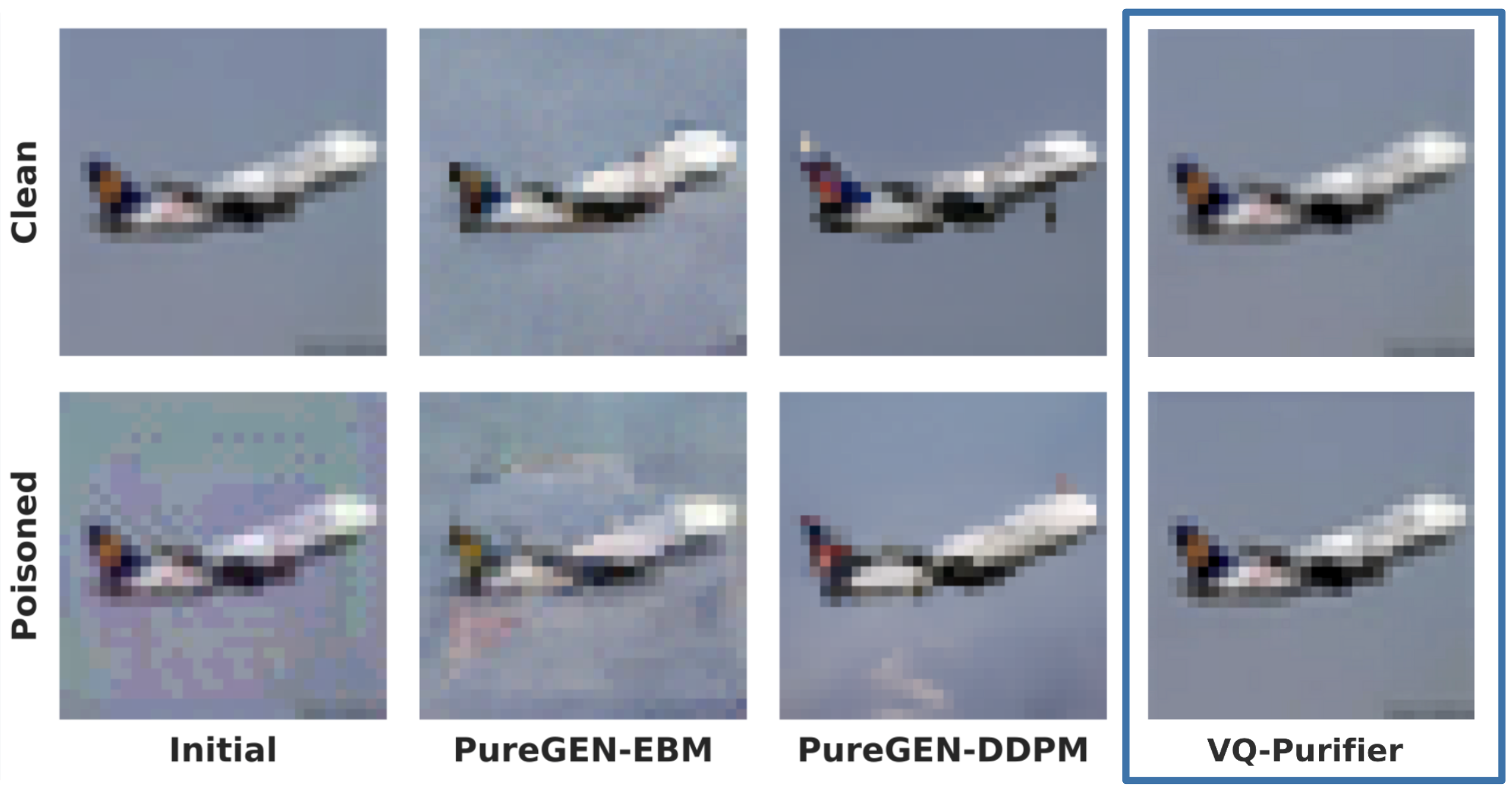}
\caption{Visual comparison. \alg\ removes triggers while maintaining image quality.}
\label{fig:comparison}
\end{figure}

\textbf{Ablation Studies:} Figure~\ref{fig:ablation}(a) shows performance scales with model size, with improvements continuing up to 50M parameters. Our main results use a 300M parameter model which achieves the best defense performance. Figure~\ref{fig:ablation}(b) demonstrates even small codebooks (K=64) achieve near-zero PSR, while larger codebooks improve reconstruction quality. Extended analysis in Appendix~\ref{app:details}.

\textbf{GAN Component Analysis:} We ablate the GAN discriminator by training a VQ-VAE-only baseline. While the discrete bottleneck alone maintains strong defense against poisoning, the reconstructions are noticeably blurry, causing downstream classifier accuracy to degrade. The adversarial loss is critical for producing crisp, high-fidelity reconstructions that preserve fine-grained semantic features needed for accurate classification. This demonstrates that the GAN component is essential not just for perceptual quality, but for maintaining the natural accuracy necessary for a practical defense.

\textbf{Trigger Analysis:} Figure~\ref{fig:comparison} shows \alg\ effectively removes both spatial patterns (NS) and imperceptible perturbations (GM). The discrete bottleneck forces representations through a finite codebook, preventing adversarial signals from propagating. High-frequency triggers get quantized to nearest clean prototypes. \alg\ achieves PSNR $>40$ dB, indicating near-lossless reconstruction (see Appendix~\ref{app:details}).
\vspace{-0.3cm}
\section{Conclusion and Future Work}

\alg\ leverages VQ-GAN's discrete bottleneck~\cite{esser2021taming} for efficient defense against data poisoning. The finite codebook filters adversarial perturbations while the GAN loss maintains natural image distribution. We achieve 0-1.64\% PSR with 91-95\% clean accuracy in a single forward pass, 50× faster than diffusion methods.

\section*{Acknowledgments}

We thank Google's TPU Research Cloud (TRC) program for providing computational resources that made this research possible.

\newpage
\bibliographystyle{IEEEtran}
\bibliography{IEEEabrv,threshoptim}

\newpage
\appendix

\section{Additional Experimental Details}
\label{app:details}

\subsection{Visual Purification Quality Analysis}

Figure~\ref{fig:comparison} in the main text shows purification results across different methods. To quantify reconstruction fidelity, we measure Peak Signal-to-Noise Ratio (PSNR) on purified images. \alg\ achieves PSNR $>40$ dB on average, indicating near-lossless reconstruction. In contrast, PureGen-EBM and PureGen-DDPM achieve PSNR of 35-38 dB, with outputs appearing slightly washed-out or distorted. This superior reconstruction quality explains why \alg\ maintains higher natural accuracy (94.61\% vs 92.92\% for PureGen-EBM) while achieving comparable poison defense.

\subsection{Qualitative Comparison of Defense Methods}

Table~\ref{tab:compare_full} provides a detailed qualitative comparison highlighting the main limitation of each prior defense and how \alg\ addresses these challenges.

\begin{table}[h]
\centering
\caption{\textbf{Qualitative Comparison of Defense Methods.}}
\label{tab:compare_full}
\begin{tabular}{p{3cm} p{5cm} p{5.5cm}}
\toprule
\textbf{Defense Method} & \textbf{Limitation / Drawback} & \textbf{\alg\ Advantage} \\
\midrule
FRIENDS & Adds small noise to all images, which can slightly degrade image quality and may not remove all poison artifacts. & Produces purified images with no added noise and near-perfect fidelity to the original (no quality degradation). \\
\addlinespace
EPIC & Eliminates suspected poisonous data from the training set, risking removal of clean data and reducing data volume for training. & Does not discard any images; all data points are retained and purified, preserving dataset size and diversity. \\
\addlinespace
PureGen (EBM/DDPM) & Requires hundreds of iterative refinement steps per image (high computational cost) and can yield somewhat blurred reconstructions. & Requires only a single forward pass for purification, enabling fast processing, and outputs high-detail images thanks to the GAN decoder. \\
\addlinespace
JPEG Compression & Simple and fast but degrades image quality by removing high-frequency components needed for classification. & Selective filtering through learned codebook preserves semantic features while removing adversarial perturbations. \\
\bottomrule
\end{tabular}
\end{table}

\subsection{Extended Model Capacity Analysis}

Beyond the main results in Figure~\ref{fig:ablation}(a), we provide additional insights into model scaling behavior. As model size grows from 5M to 50M parameters, natural accuracy improves from approximately 88\% to 92\%, while PSR remains near zero across all model sizes. The largest models (up to 300M parameters) show diminishing returns, with only marginal accuracy gains beyond 50M parameters. This suggests that defense effectiveness is primarily determined by the discrete bottleneck architecture rather than model capacity, while larger models mainly improve reconstruction quality.

\subsection{Codebook Usage Analysis}

Figure~\ref{fig:ablation}(b) examines varying codebook size $K$ for a fixed model architecture. We observe that increasing from $K=64$ to $K=512$ yields improvements in natural accuracy as a larger dictionary allows finer image details to be represented. However, further enlarging to $K=2048$ gives only minor gains, and code usage saturates around 500-600 codes. This implies an intrinsic limit on how much detail the CIFAR-10 data distribution requires. Importantly, even the smallest codebook ($K=64$) already drives PSR to effectively 0\%, highlighting that quantization is the primary factor disabling poisons, while codebook size mainly affects reconstruction quality.

\section{Implementation Details}
\label{app:implementation}

\subsection{Training Hyperparameters}

We train all \alg\ models using the following hyperparameters:
\begin{itemize}
    \item Optimizer: Adam with learning rate $4 \times 10^{-4}$
    \item Batch size: 256
    \item Training epochs: 100
    \item Commitment loss weight: $\beta = 0.25$
    \item GAN loss weight: $\lambda = 0.1$
    \item Discriminator architecture: PatchGAN with 3 layers
    \item Image resolution: 32×32 (CIFAR-10)
    \item Latent spatial resolution: 8×8
    \item Latent channels: $d = 256$
\end{itemize}

\subsection{Architecture Details}

The encoder consists of 4 residual blocks with stride-2 convolutions to downsample from 32×32 to 8×8. The decoder mirrors this structure with transposed convolutions. The discriminator follows the PatchGAN architecture with spectral normalization. All models use GroupNorm instead of BatchNorm for better stability during adversarial training.

\end{document}